\pdfoutput=1
\documentclass[10pt,twocolumn,letterpaper]{article}

\usepackage[pagenumbers]{iccv} % To force page numbers, e.g. for an arXiv version

\usepackage[accsupp]{axessibility}
\usepackage{amsmath}
\usepackage{afterpage}
\usepackage{amssymb}
\usepackage{booktabs}
\usepackage{pifont}
\usepackage{color}
\usepackage{bm}
\usepackage{multirow}
\usepackage{soul}
\usepackage{bbm}
\usepackage{graphicx}
\usepackage{xcolor}
\usepackage{array}

\newcommand{\xmark}{\ding{55}}
\definecolor{citecolor}{HTML}{0071bc}
\usepackage[pagebackref=true,breaklinks=true,colorlinks,citecolor=citecolor,bookmarks=false]{hyperref}

\def\paperID{11} % *** Enter the Paper ID here
\def\confName{ICCV}
\def\confYear{2025}

\title{Toward Scalable Video Narration: A Training-free Approach \\ Using Multimodal Large Language Models}

\author{
Tz-Ying Wu\thanks{These authors contributed equally to this work.} \quad Tahani Trigui$^{\ast}$ \quad Sharath Nittur Sridhar \quad Anand Bodas \quad Subarna Tripathi
\\
Intel
\\ \tt\footnotesize \{tz-ying.wu,tahani.trigui,sharath.nittur.sridhar,anand.v.bodas,subarna.tripathi\}@intel.com
}

\begin{document}

\maketitle

\begin{abstract}
In this paper, we introduce {\bf VideoNarrator}, a novel training-free pipeline designed to generate dense video captions that offer a structured snapshot of video content. These captions offer detailed narrations with precise timestamps, capturing the nuances present in each segment of the video. Despite advancements in multimodal large language models (MLLMs) for video comprehension, these models often struggle with temporally aligned narrations and tend to hallucinate, particularly in unfamiliar scenarios. {\bf VideoNarrator} addresses these challenges by leveraging a flexible pipeline where off-the-shelf MLLMs and visual-language models (VLMs) can function as caption generators, context providers, or caption verifiers. Our experimental results demonstrate that the synergistic interaction of these components significantly enhances the quality and accuracy of video narrations, effectively reducing hallucinations and improving temporal alignment.
This structured approach not only enhances video understanding but also facilitates downstream tasks such as video summarization and video question answering, and can be potentially extended for advertising and marketing applications.
\end{abstract}
  
\section{Introduction}

Video is a multidimensional signal, encapsulating the dynamic scenes and complex visual details across spatial and temporal dimensions. This characteristic makes it an influential medium for recording, communication, entertainment, and advertising.
Despite containing vast amounts of information, videos are inherently low-level and demand substantial storage space. Moreover, retrieving specific information from very long videos in response to a query can be challenging and inefficient if done frequently. It is therefore essential to extract the core content of the video and preserve it in a more concise format, such as dense video captioning (DVC), where narrations are provided with their timestamps, such as ``52.2s - 74.4s the person is then spreading mayonnaise on the bread."
This creates a structured snapshot of the video, capturing the scene semantics and dynamics within each video segment, that can be potentially extended for downstream applications in advertising and marketing, e.g., understanding visual advertisements~\cite{understanding_of_visual_ads, Ye_2018_ECCV, MM-AU2023}, and analyzing user or influencer videos for targeted marketing~\cite{agrawal2025optimizing} in several domains including ad-personalization, retail~\cite{video_understanding_retail}, and e-commerce~\cite{video_understanding_ecommerce}.

\begin{figure}[t!]
    \centering
    \includegraphics[width=.95\linewidth]{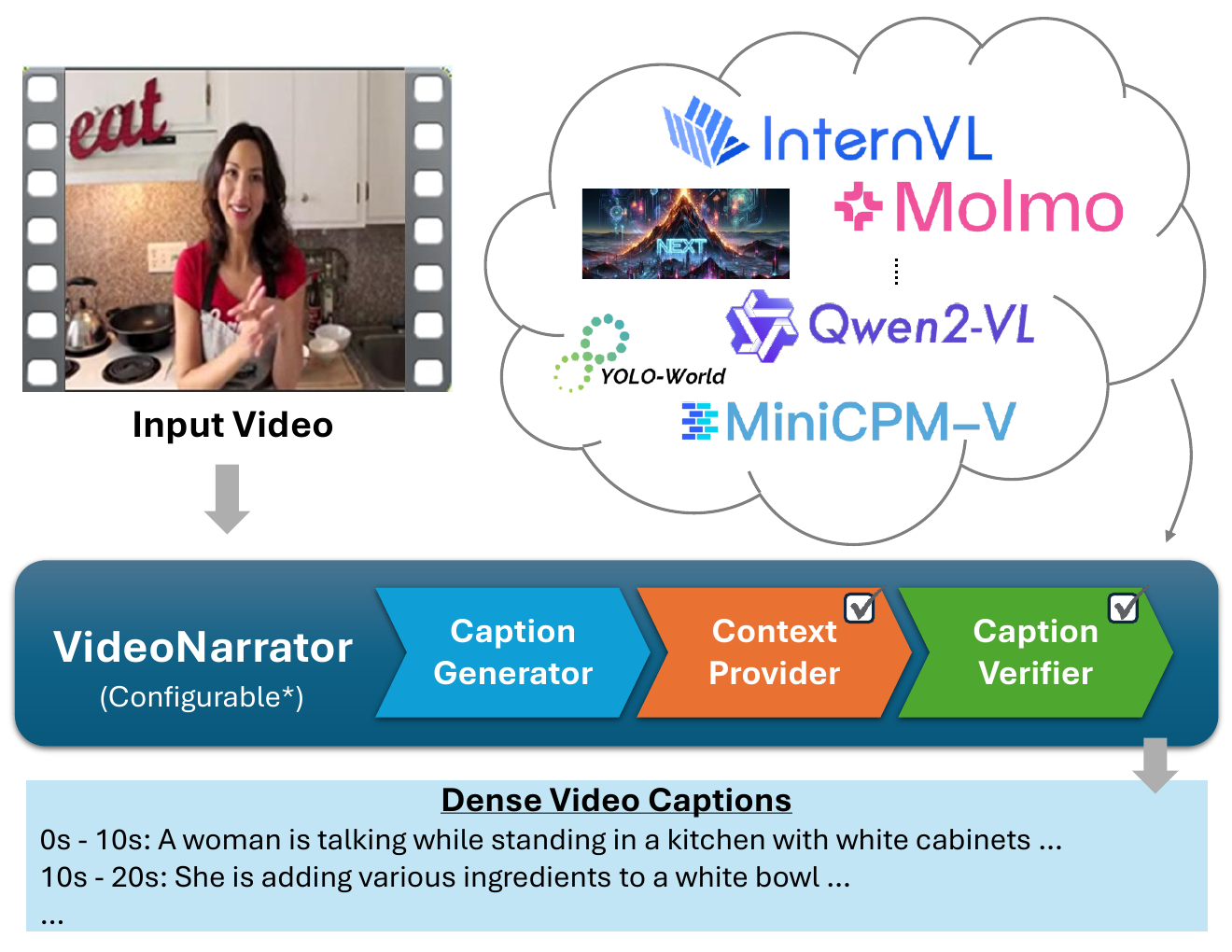}
    \vspace{-5pt}
    \caption{{\bf VideoNarrator} is a {\it training-free} and configurable pipeline harnessing the power of off-the-shelf MLLMs and VLMs for dense video captioning, establishing a scalable solution for real-world video understanding tasks.}
    \label{fig:teaser}
\end{figure}

While videos are widely accessible from multiple sources, DVC annotations are costly to obtain and thus sparsely available, limiting the training and evaluation scope in prior DVC research~\cite{denseVidCapSurvey2025}.
In contrast, the recent advances that bridge visual and language domains present a new opportunity: generate video narrations for {\it any} video using common knowledge acquired from a broader range of datasets. For example, a general purpose multimodal large language model (MLLM)~\cite{MLLM_survey2023} can be guided to describe the content at regular intervals (for every $S$ seconds). Although promising, this approach remains underexplored. Since these models are not specifically tailored to the target video, the resulting video narrations may not always be reliable and could include inaccuracies or hallucinations.

To address this, we propose {\bf VideoNarrator}, a {\it training-free} pipeline for reducing hallucinations and improving the quality of DVC. This framework employs a modular design, leveraging existing MLLMs and visual-language models (VLMs) to serve as {\it caption generators}, {\it context providers}, or {\it caption verifiers}, where each component plays a distinct role: generating narrations, supplying scene context, and detecting hallucinated captions, respectively. For example, an object detector~\cite{YOLOWorldCheng2024, object_detector_survey_2024} can be a {\it context provider}, offering rich semantics about the scene to supplement a {\it caption generator} for crafting more relevant captions, while a {\it caption verifier} can be utilized to identify and eliminate inaccuracies. The synergy of these roles improves the accuracy and relevance of the captions.

\begin{figure}[t!]
    \centering
        \centering
        \includegraphics[width=.93\linewidth]{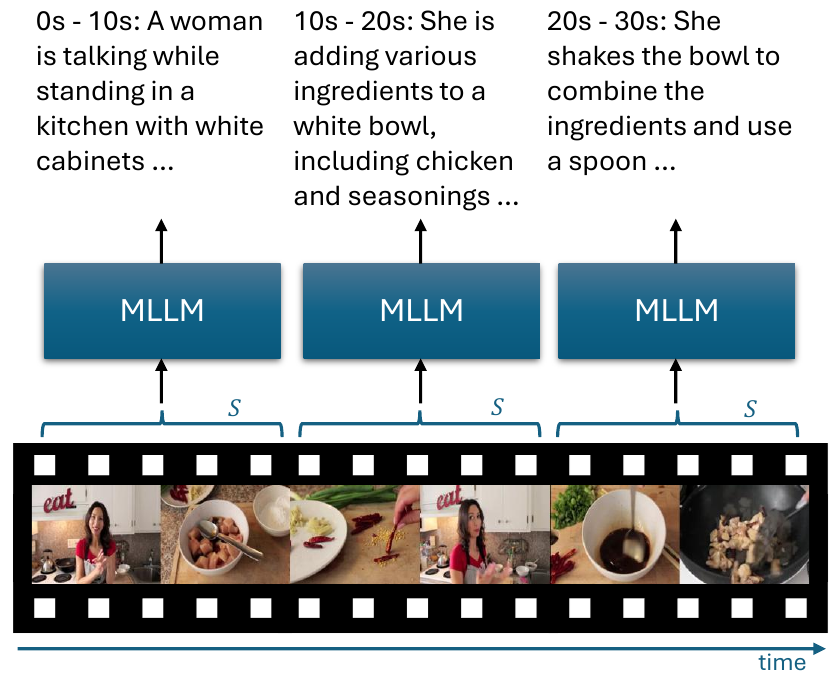}
        \vspace{-5pt}
        \caption{Dense video captioning with MLLMs. Videos are segmented into chunks with uniform intervals (i.e., $S$ seconds), and the MLLM generates the caption for each segment individually.}
        \label{fig:mllm}
\end{figure}

\begin{figure*}[t!]
        \centering
        \includegraphics[width=0.84\linewidth]{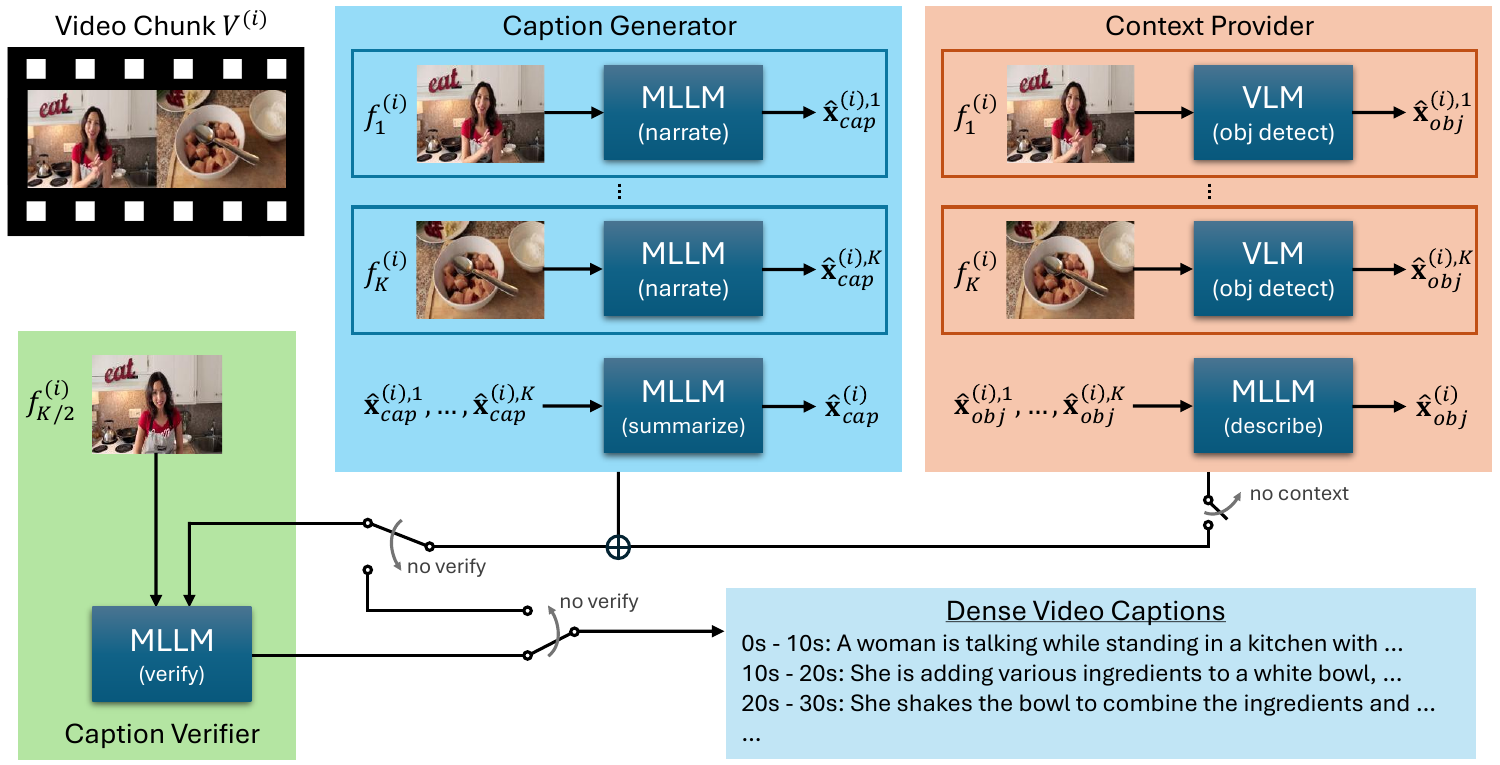}
        % \vspace{-5pt}
        \caption{The {\bf VideoNarrator} pipeline includes MLLM and VLM modules functioned for different purposes: {\it caption generator}, {\it context provider}, and {\it caption verifier}. It is a {\it training-free} and configurable framework. The video in the example is sourced from \cite{Youcook2ZhXuCoAAAI18}.}
        \label{fig:pipeline}  
\end{figure*}

For quantitative assessment of these components, we introduce an evaluation protocol that measures the quality of video captions through a multiple-choice question answering (MCQ) task using the Video-MME~\cite{videoMMEfu2024video} dataset, which comprises a wide range of questions associated with diverse videos.
Extensive experiments demonstrate that by integrating these roles, {\bf VideoNarrator} effectively enhances the reliability of video descriptions, offering a scalable solution for generating high-quality narrations without the necessity for extensive training tailored for specific use cases.

In summary, the paper makes the following contribution:
\begin{itemize}
    \item We propose \textbf{VideoNarrator}, a {\it training-free} DVC framework that enhances caption quality and reliability through modular integration of existing MLLMs and VLMs.
    \item We enhance caption accuracy and relevance by leveraging semantic scene information and hallucination detection, reducing common errors in video narration.
    \item We present a new evaluation protocol based on multiple-choice question answering using the Video-MME dataset, offering robust quantitative assessment of DVC performance.
\end{itemize}

\section{Related Work}
\subsection{Dense Video Captioning}
Dense video captioning (DVC) can be thought of as the combination of event localization and event captioning~\cite{krishna2017dense,wang2021end}. 
DVC has been regarded as highly useful in applications such as large-scale video search and indexing. 
With the advent of LLMs and GenAI, DVC is turning out to be an extremely useful component for video question-answering and long video summarization as well. 
An initial line of work~\cite{krishna2017dense, MDVC_Iashin_2020, BMT_Iashin_2020, Wang2020echr} that approaches DVC, follows a two-stage process; temporal localization stage followed by event captioning stage. Recent work, on the other hand, looked at joint optimization for captioning and localization~\cite{deng2021sketch,zhou2018end}. Please refer to~\cite{denseVidCapSurvey2025} for an in-depth study of several methods, evaluations and datasets for the DVC tasks. 

\subsection{MLLMs for Video Understanding}
LLMs integrated with video as input modality introduced a new paradigm in video understanding~\cite{maaz2023video,lin2023video,zhang2023video}. These models are equipped with multimodal reasoning power, and enable effective ways of interacting with videos with free-form textual prompts. Thanks to the convenience of use, such video-LLM models are becoming pervasive in several domains and use cases. The survey paper~\cite{tang2024videounderstandingLL} provides a great deal of insights on different video-LLM models, their primary use cases and usability. Most of these Video-LLM models leverage open-sourced LLMs like LlaMA~\cite{touvron2023llamaopenefficientfoundation} or Vicuna~\cite{chiang2023vicuna} as the backbone. Recent video-LLMs have become omnipresent for video applications including video classification to video question-answering~\cite{li2024llama,song2024moviechat}, bridging the gap between human-level performance and previously existing discriminative video models in terms of reasoning capabilities. However, video-LLMs tailored for dense video captioning are comparatively underexplored~\cite{ren2024timechat,huang2024lita}, possibly because the limitedly available DVC datasets for supervised training. Alternatively, we explore the potential of using general purpose MLLMs for tackling the task of DVC without further training. We leverage the common knowledge acquired from diverse visual-language datasets with supplementary information from VLMs, aiming for a more scalable solution for real-world video narration. Note that while we focus on general purpose MLLMs in this work, the {\bf VideoNarator} pipeline is general and can be applied to video-LLMs for DVC~\cite{ren2024timechat,huang2024lita} as well.

\section{Dense Video Captioning with MLLMs}

Given a video $V$, the task of dense video captioning (DVC) involves generating a sequence of narrations, each paired with a corresponding temporal segment, i.e., $\{(t_{st}^{(i)},t_{end}^{(i)}, {\bf x}_{cap}^{(i)})\}$, where ${\bf x}_{cap}^{(i)}$ denotes the textual description of the event occurring between timestamps $t_{st}^{(i)}$ and $t_{end}^{(i)}$, and $i$ the index of the temporal segment.
The task is challenging because the caption ${\bf x}_{cap}^{(i)}$ is only valid when the temporal localization $(t_{st}^{(i)},t_{end}^{(i)})$ of the event is precise, since video content changes from time to time.
The conventional approach for tackling DVC is to train a model with supervised DVC datasets in an end-to-end fashion~\cite{wang2021end}. However, acquiring DVC annotations is labor-intensive, which constraints the availability of such supervised data, limiting the scope of training and evaluation. 

In this paper, we investigate the potential of utilizing general purpose multimodal large language models (MLLMs) that support vision-language understanding for tackling DVC.
MLLMs leverage the superb capability of large language models (LLMs) in context reasoning, by aligning data from other modalities to the language domain.
An MLLM consumes two types of inputs: a visual input (an image or a video) and a text prompt, which can be a question about the given video or a specific instruction (e.g., ``describe the image").
The features of the visual input are extracted by a visual encoder, projected into the token space of the LLM, and jointly interpreted alongside the textual tokens.
This allows the interaction between the two modalities, and enables generating descriptions for {\it arbitrary} visual content.

While MLLMs possess broad knowledge acquired from diverse datasets, they are not explicitly fine-tuned for DVC, often resulting in undesired performance in temporal localization. These models, however, demonstrate strong capabilities in observing and describing content in images or short video segments.
To capitalize on this, rather than prompting a MLLM to generate dense captions directly, we chunk the video into several uniform segments (i.e., every $S$ seconds) and instruct the model to describe each video chunk $V^{(i)}$ independently as depicted in Figure~\ref{fig:mllm}, similar to~\cite{islam2024video}. This strategy naturally yields captions with accurate and readily available temporal boundaries, where each caption spans from $t_{st}^{(i)}$ to $t_{end}^{(i)}=t_{st}^{(i)}+S$.
Nevertheless, the captions are susceptible to contain hallucinated factual elements especially when the input video is from an unseen scenario. We hypothesize that these inaccuracies can be mitigated by a workflow that integrates the power of different models on content generation, context extraction, and verification. In the next section, we explore such a hypothesis by introducing {\bf VideoNarrator}, a {\it training-free} pipeline for tackling DVC using MLLMs, and discuss different roles within the pipeline. 

\section{VideoNarrator}

The aforementioned DVC with MLLMs approach enhances the scalability of video narration for {\it in-the-wild} videos. However, the resulting narrations are still prone to include hallucinations.
To address this and improve caption quality, we propose a {\it training-free} pipeline, {\bf VideoNarrator}, that embodies the {\it together-makes-better} hypothesis, harnessing the complementary strength of multiple models, each dedicated to a specific function detailed below.

\paragraph{Caption Generator:} A module that employs a MLLM to provide the initial caption prediction $\hat{\bf x}_{cap}^{(i)}$ for each video segment $V^{(i)}$, which is achieved by prompting the model with the instruction, {\it ``Describe the activities and events captured in the image. Provide a detailed description of what is happening,"} where $K$ frames of the video segment are sampled as the visual input, i.e., $f_1^{(i)},...,f_K^{(i)}$. We generate the narration $\hat{\bf x}_{cap}^{(i),j}$ for individual frames $f_j^{(i)}$ and prompt the model again to summarize the captions across the frames within the chunk, as illustrated in the {\it caption generator} block of Figure~\ref{fig:pipeline}.

\paragraph{Context Provider:}
A module for extracting scene semantics, such as an object detector, enriching the initial predictions from the {\it caption generator} with more detailed context.
In this work, we utilize a VLM-based object detector, YOLO-World~\cite{YOLOWorldCheng2024} to detect the most visible objects in each sampled frame $f_j^{(i)}$ of the video chunk $V^{(i)}$. The detected objects $\hat{\bf x}_{obj}^{(i),1},...,\hat{\bf x}_{obj}^{(i),K}$ are then passed into a MLLM to acquire the object description $\hat{\bf x}_{obj}^{(i)}$, which is then appended to the initial narration, i.e., $\hat{\bf x}_{cap}^{(i)}\oplus\hat{\bf x}_{obj}^{(i)}$, where $\oplus$ denotes sequence concatenation. The overall process is summarized in the orange block of Figure~\ref{fig:pipeline}.

\paragraph{Caption Verifier:} A module harnessing a MLLM to verify the narrations predicted by the previous steps, depicted in the green block of Figure~\ref{fig:pipeline}. Unlike {\it caption generator}, the model does not need to create content, but focuses on checking the correctness of the given caption with respect to the visual input. Specifically, we prepend the caption from {\it caption generator} (and {\it context provider}) to the instruction, {\it ``Does this accurately describe the given content? Simply answer Yes/No,"} and prompt the model with the middle frame $f_{K/2}^{(i)}$ of the video chunk $V^{(i)}$. The captions receiving a ``No" in the answer are filtered out for error prevention, and only the rest are preserved.

All the components introduced above assemble the {\bf VideoNarrator} pipeline, where the modular design offers seamless integration of off-the-shelf MLLMs and VLMs, each fulfilling specialized roles, thereby contributing to generate more reliable and relevant video narrations. This {\it training-free} and configurable characteristic also enhances the scalability of this approach, making it adaptable to a wide range of video content without the supervisions.

\section{Experiments}
In this section, we present the empirical analysis of the {\bf VideoNarrator} performance.

\begin{figure}[t!]
    \centering
    \includegraphics[width=\linewidth]{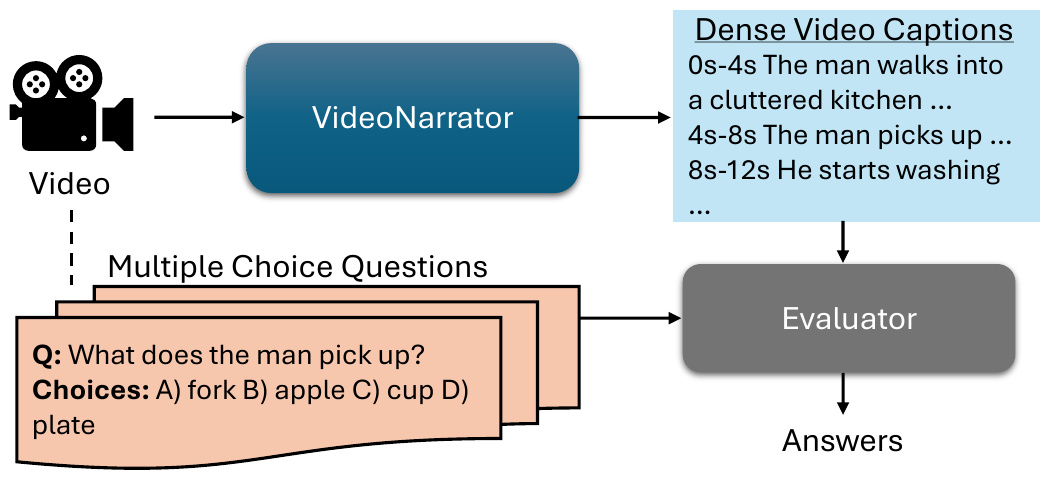}
    \captionof{figure}{Evaluation protocol based on multiple choice question (MCQ) answering. The evaluator takes the dense captions produced by {\bf VideonNarrator} to answer the corresponding MCQs.}
    \label{fig:eval_flow}
\end{figure}

\subsection{Evaluation Protocol}\label{sec:evaluation}
Traditional evaluation of DVC primarily relies on the direct comparisons between the predicted captions and the human-annotated ground truth. However, this confines the evaluation to datasets containing DVC-specific annotations, but does not scale to other video domains. Moreover, standard metrics may fall short when assessing longer and more complex captions produced by MLLMs, due to the constraints such as $n$-gram matching~\cite{vedantam2015cider,banerjee_meteor_2005} and context length limitation in the feature extractors~\cite{zhang_bertscore_2020}.
As an alternative, we assess the caption quality through the interaction to the video content via question answering.
Specifically, we introduce an evaluation protocol based on multiple choice question (MCQ) answering to measure the correctness and informativeness of DVC outputs.
We adopt a subset of VideoMME~\cite{videoMMEfu2024video} as the evaluation dataset, which contains videos spanning across different domains and subcategories, and employ {\tt Llama3.1-8B-Instruct}~\cite{grattafiori2024llama} as the evaluator. For each test video, the model answers the associated MCQs based solely on the dense captions produced by the {\bf VideoNarrator} pipeline. Since the evaluator lacks direct access to the video, the captions must convey accurate and semantically rich information for answering the MCQs correctly. We report the accuracy of the MCQ answering as the primary metric for evaluating different system configurations.

\subsection{Settings}\label{sec:settings}
{\bf VideoNarrator} is a general pipeline that supports off-the-shelf MLLMs. We consider the following state-of-the-art MLLMs for the experiments:
\begin{itemize}
    \item {\tt InternVL2-1B/4B}~\cite{internvl2_chen2024far}: Employing the powerful InternViT~\cite{chen2024internvl} model as the visual encoder, based on large-scale contrastive pretraining,
    supporting a wide range of multimodal comprehension tasks, such as document and chart analysis, OCR, and scene text understanding. We consider their lightweight versions here.
    \item {\tt Molmo-7B-D-0924}~\cite{molmo2024}: An open-weight and open-data MLLM pretrained with highly detailed image captions and object pointing and counting data~\cite{molmo2024}. It is trained end-to-end and does not require synthetic data distilled from other close-source MLLMs. 
    \item {\tt Qwen2-VL-7B-Instruct}~\cite{qwen2vlwang2024}: Featuring multimodal perception with dynamic resolutions and aspect ratios, multi-lingual OCR, long-form video understanding, and reasoning across multiple images.
    \item {\tt MiniCPM-V-2.6}~\cite{hu2024minicpm}:
    Delivering efficient output without compromising performance, achieving superior performance than GPT-4v, while being lightweight (with 8B model parameters) and deployable to edge devices.
    \item {\tt Llama3-Llava-next-8B}~\cite{liu2024llavanext}: A MLLM based on {\tt Meta-Llama3}, which excels in high-resolution visual reasoning and supports multi-image/video inference.
\end{itemize}
In all cases, the same MLLM is adopted throughout the pipeline, serving as the MLLM in all the components enabled. For video chunking, $S$ and $K$ are set to 10 and 2, respectively.
Note that we adopt the YOLO-World~\cite{YOLOWorldCheng2024} object detector in {\it context provider} irrespective of the MLLM choice.

\subsection{Main Results}

We ablate the effect of different components in the {\bf VideoNarrator} pipeline with state-of-the-art MLLMs.  
Table~\ref{tab:w_obj_det} reports the impact of incorporating object semantics as contextual information via the {\it context provider}, while the {\it caption verifier} remains disabled. The results show accuracy improvements across models, except for {\tt Molmo-7B} whose performance remains unchanged. This can be attributed to {\tt Molmo} being more proficient at object grounding, stemming from its training on the PixMo~\cite{molmo2024} dataset. The initial narration generated by {\tt Molmo} could already be rich in semantics, making the effect of additional object detection marginal. However, for the remaining models, integrating object-level context appears to be an effective complement to the {\it caption generator}.

Similarly, we investigate the effect of enabling the {\it caption verifier} in isolation, without the {\it context provider}. Contrary to the previous observations, results shown in Table~\ref{tab:w_verifier} suggest that integrating the {\it caption verifier} alone do not yield clear improvements over the baseline. This is likely due to the inherently passive nature of the verification process. It does not fundamentally change the caption content but merely filters out the potential errors. When the filtering is too aggressive, it may also result in the loss of useful information.

\begin{table}[t!]
    \centering
    \setlength{\tabcolsep}{5pt}
    \resizebox{\linewidth}{!}{
    \begin{tabular}{l|c|c}
        \toprule
        Model & + Obj. Context & Accuracy (\%)\\ \midrule
        InternVL2-1B~\cite{internvl2_chen2024far} & \xmark & 40.00 \\
        InternVL2-1B~\cite{internvl2_chen2024far} & \checkmark & 42.22 \\ \midrule
        InternVL2-4B~\cite{internvl2_chen2024far} & \xmark & 40.00 \\
        InternVL2-4B~\cite{internvl2_chen2024far} & \checkmark & 48.89 \\ \midrule
        Molmo-7B-D-0924~\cite{molmo2024} & \xmark & 44.44 \\
        Molmo-7B-D-0924~\cite{molmo2024} & \checkmark & 44.44 \\ \midrule
        Qwen2-VL-7B-Instruct~\cite{qwen2vlwang2024} & \xmark & 40.00 \\
        Qwen2-VL-7B-Instruct~\cite{qwen2vlwang2024} & \checkmark & 44.44 \\ \midrule
        MiniCPM-V-2.6~\cite{hu2024minicpm} & \xmark & 44.44 \\
        MiniCPM-V-2.6~\cite{hu2024minicpm} & \checkmark & 46.67 \\
        \bottomrule        
    \end{tabular}
    }
    \captionof{table}{Effect of providing object information as context to the video narration generator. {\it Caption verifier} is not enabled here.}
    \label{tab:w_obj_det}
\end{table}

\begin{table}[t!]
    \centering
    \setlength{\tabcolsep}{7pt}
    \resizebox{\linewidth}{!}{
    \begin{tabular}{l|c|c}
        \toprule
        Model & + Verifier & Accuracy (\%)\\ \midrule
        InternVL2-4B~\cite{internvl2_chen2024far} & \xmark & 40.00 \\
        InternVL2-4B~\cite{internvl2_chen2024far} & \checkmark & 46.67 \\ \midrule
        Molmo-7B-D-0924~\cite{molmo2024} & \xmark & 44.44 \\
        Molmo-7B-D-0924~\cite{molmo2024} & \checkmark & 44.44 \\ \midrule
        Qwen2-VL-7B-Instruct~\cite{qwen2vlwang2024} & \xmark & 40.00 \\
        Qwen2-VL-7B-Instruct~\cite{qwen2vlwang2024} & \checkmark & 40.00 \\ \midrule
        MiniCPM-V-2.6~\cite{hu2024minicpm} & \xmark & 44.44 \\
        MiniCPM-V-2.6~\cite{hu2024minicpm} & \checkmark & 42.22 \\ \midrule
        LLama3-Llava-next-8B~\cite{liu2024llavanext} & \xmark & 44.44 \\
        LLama3-Llava-next-8B~\cite{liu2024llavanext} & \checkmark & 44.44 \\
        \bottomrule        
    \end{tabular}
    }
    \caption{Effect of verifying the generated captions with the same model. {\it Context provider} is not enabled here.}
    \label{tab:w_verifier}
\end{table}

\begin{table}[t!]
    \centering
    \setlength{\tabcolsep}{2pt}
    \resizebox{\linewidth}{!}{
    \begin{tabular}{l|c|c}
        \toprule
        Model & + Obj. Context + Verifier & Accuracy (\%)\\ \midrule
        InternVL2-4B~\cite{internvl2_chen2024far} & \xmark & 40.00 \\
        InternVL2-4B~\cite{internvl2_chen2024far} & \checkmark & 46.67 \\ \midrule
        Molmo-7B-D-0924~\cite{molmo2024} & \xmark & 44.44 \\
        Molmo-7B-D-0924~\cite{molmo2024} & \checkmark & 48.89 \\ \midrule
        Qwen2-VL-7B-Instruct~\cite{qwen2vlwang2024} & \xmark & 40.00 \\
        Qwen2-VL-7B-Instruct~\cite{qwen2vlwang2024} & \checkmark &  46.67 \\ \midrule
        MiniCPM-V-2.6~\cite{hu2024minicpm} & \xmark & 44.44 \\
        MiniCPM-V-2.6~\cite{hu2024minicpm} & \checkmark & 42.22 \\ \midrule 
        LLama3-Llava-next-8B~\cite{liu2024llavanext} & \xmark & 44.44 \\
        LLama3-Llava-next-8B~\cite{liu2024llavanext} & \checkmark & 53.33 \\     
        \bottomrule        
    \end{tabular}
    }
    \caption{Effect of including both {\it context provider} and {\it caption verifier} in the pipeline.}
    \label{tab:full}
\end{table}

\begin{figure*}
    \centering
    \includegraphics[width=\linewidth]{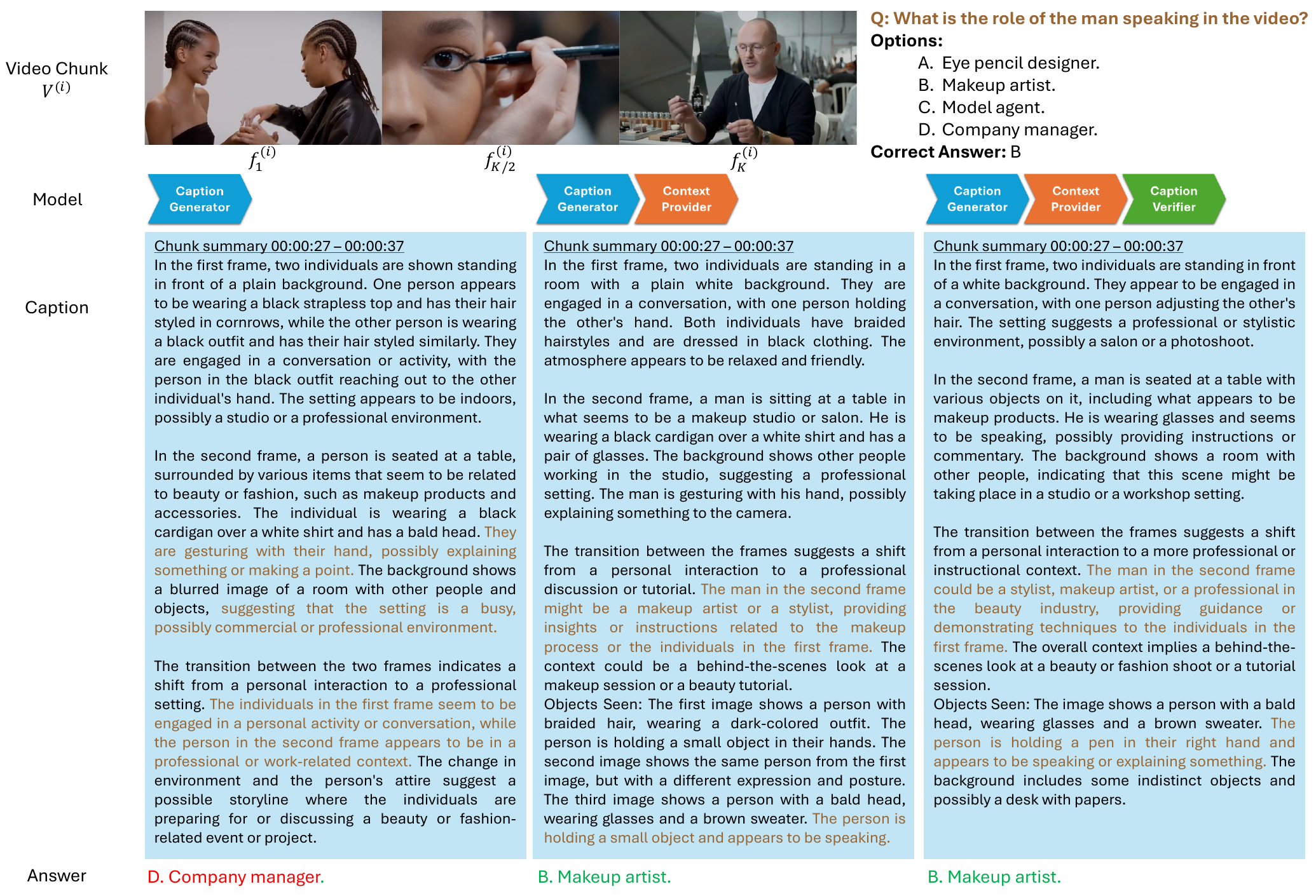}
    \caption{Visualization of {\bf VideoNarrator} outputs and their corresponding answers to the MCQ (shown on the upper right) of different configurations: (left) {\it Caption generator} only. (middle) {\it Caption generator} + {\it context provider}. (right) {\it Caption generator} + {\it context provider} + {\it caption verifier}, where the first one suggested an incorrect answer (red) to the given MCQ, and only the latter two chose the correct answer (green). Content related to the question is marked in brown.}
    \label{fig:visualize}
\end{figure*}

Finally, we examine the full pipeline of the {\it VideoNarrator}, enabling both the {\it context provider} and the {\it caption verifier}, as shown in Table~\ref{tab:full}. The results demonstrate notable gain over the baseline across MLLMs other than {\tt MiniCPM-V}. We notice that the accuracy for this model is also dropped when only the {\it caption verifier} is enabled. It is possible that the model acts overly stringent as the verifier, eliminating too much information.
Another interesting observation is that the performance of {\tt Molmo} was not improved when the two components are added individually, while the accuracy increases $4.45\%$ when both modules are enabled, probability because the object description provided by the {\it context provider} is mixed with correct and incorrect information. Without the verification process, the correct ones cannot stand out.
Similar observation is seen on {\tt Qwen2-VL} and {\tt Llama3-Llava-next}, where the accuracy improves over their counter part without the {\it context provider}.
These results suggest that the {\it context provider} and the {\it caption verifier} work complementarily to each other, they are most effective when they work together. While the former strives to deliver meaningful object-aware context to enrich understanding, the latter dedicates to sift the truth from the noise, retaining only verifiable content while systematically discarding what is false or misleading, which enhances the video narration quality.

\subsection{Additional Analysis}
We extend our ablation study to examine additional factors influencing performance.
\paragraph{Chunk size and frame rate.}
Table~\ref{tab:chunk_size} and Table~\ref{tab:nframe} analyze the effect of varying the chunk size $S$ and the number of frames per chunk $K$, respectively. Increasing the chunk size provides a broader temporal context within each segment, while using more frames per chunk enhances the video's temporal resolution. Both factors contribute to improve contextual reasoning and thereby improve the overall performance.  

\paragraph{Model quantization.}
Table~\ref{tab:quantize} ablates the effect of quantization. Note that both models here represent the complete {\bf VideoNarrator} configuration, integrating both the object context and verification, and the multi-image inference is not adopted as the {\tt Qwen2-VL} model in other tables.
The result show that quantization using AWQ does not lead to significant performance drop, especially when compared to the gains introduced by the {\bf VideoNarrator} components.

\begin{figure}[t!]
    \centering
    \includegraphics[width=0.7\linewidth]{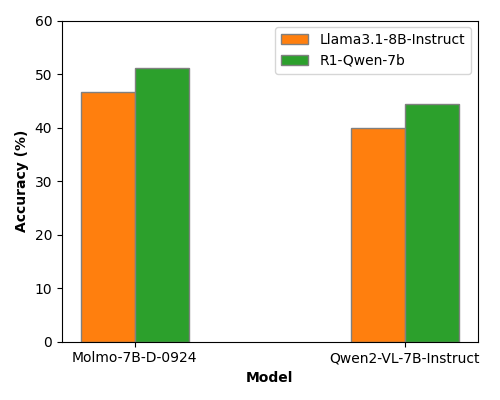}
\caption{Evaluate the MCQs with different evaluators, {\tt Llama3.1-8B-Instruct} and {\tt R1-Qwen-7b}, using the protocol introduced in section~\ref{sec:evaluation} with a subset of the evaluation data. The trend between {\tt Molmo} and {\tt Qwen2-VL} remains the same irrespective to the evaluator choice.}
    \label{fig:evaluator}
\end{figure}

\paragraph{Evaluator choice.}
Figure~\ref{fig:evaluator} illustrates a comparative analysis of two evaluators, {\tt Llama3.1-8B-Instruct}~\cite{grattafiori2024llama} and {\tt R1-Qwen-7b}~\cite{deepseekr1}, applied to a subset of the evaluation data.
The MCQ accuracy computed via the evaluation flow of Figure~\ref{fig:eval_flow} reflects the evaluator's ability in referring the dense captions and retrieving information relevant to the questions. Notably, {\tt R1-Qwen} consistently yields higher accuracy scores than {\tt Llama3.1} for the same model outputs. However, the relative performance trend between two DVC models, {\bf VideoNarrator} with {\tt Molmo} and {\tt Qwen2-VL} respectively, remain consistent across evaluators. This stability demonstrates that the proposed evaluation protocol is a robust and reliable quantitative measure for assessing DVC systems.

\subsection{Qualitative Results}
We further visualize the captions and the associated MCQ answers produced by different configurations of {\bf VideoNarrator} using {\tt Qwen2-VL-7B-Instruct} in Figure~\ref{fig:visualize}, where the sentences that might be connected to the model's answer are marked in brown.
Note that we only show the chunk summary related to the question (i.e., between 27s to 37s of the video), and omit the captions for other time segments for saving the space, while the dense captions of the whole video is accessible for the evaluator for contextual reasoning.
The left column shows the caption generated by the vanilla MLLM, where the narration are more generic and the description about the commercial setting might be the reason that the model answers ``company manager" instead of the correct answer ``makeup artist." The middle column corresponds to the model incorporating object context, while the right column reflects the model enhanced with both object context and a verifier. Both produce more specific scene details and generate correct answers to the given MCQ.

\begin{table}[t!]
    \centering
    \setlength{\tabcolsep}{9pt}
    \resizebox{0.85\linewidth}{!}{
    \begin{tabular}{lcc}
        \toprule
        Model & Chunk Size & Accuracy (\%) \\ \midrule
        InternVL2-4B~\cite{internvl2_chen2024far} & 5 & 42.22 \\
        InternVL2-4B~\cite{internvl2_chen2024far} & 10 & 46.67 \\
        \bottomrule
    \end{tabular}
    }
    \caption{Ablations on the number of frames per chunk.}
    \label{tab:nframe}
\end{table}

\begin{table}[t!]
    \centering
    \setlength{\tabcolsep}{5pt}
    \resizebox{0.9\linewidth}{!}{
    \begin{tabular}{lcc}
        \toprule
        Model & \# of frames per chunk & Accuracy (\%) \\ \midrule
        InternVL2-4B~\cite{internvl2_chen2024far} & 2 & 46.67 \\
        InternVL2-4B~\cite{internvl2_chen2024far} & 4 & 55.56 \\
        \bottomrule
    \end{tabular}
    }
    \caption{Ablations on the chunk size.}
    \label{tab:chunk_size}
\end{table}

\begin{table}[t!]
    \centering
    \setlength{\tabcolsep}{5pt}
    \resizebox{0.9\linewidth}{!}{
    \begin{tabular}{lcc}
        \toprule
        Model & Multi-Image & Accuracy (\%) \\ \midrule
        Qwen2-VL-7B-Instruct~\cite{qwen2vlwang2024} & \checkmark & 44.44 \\
        Qwen2-VL-7B-Instruct-AWQ~\cite{qwen2vlwang2024} & \checkmark & 42.22 \\
        \bottomrule
    \end{tabular}
    }
    \caption{Performance with and without quantization using the full pipeline, where both models are not using multi-image inference.}
    \label{tab:quantize}
\end{table}

\section{Conclusions and Discussions}

We introduced {\bf VideoNarrator}, a training-free pipeline for generating dense video captions offering a structured snapshot of video content. {\bf VideoNarrator} leverages off-the-shelf tools such as MLLMs and VLMs to avail functionalities such as caption generation, context augmentation and caption verification in a unified plug-and-play mechanism. Through extensive evaluations, we show that this structured approach improves the quality and accuracy of video narrations across model selections, with improved temporal alignment and reduced hallucinations. 
We also proposed an evaluation protocol that measures the quality of video captions through a multiple-choice question (MCQ) answering task using the Video-MME~\cite{videoMMEfu2024video}, which comprises a wide range of questions associated with diverse videos.
In the experiments, we considered the MLLM component to be fixed throughout the entire pipeline, as described in section~\ref{sec:settings}. However, we encourage future research to explore the options of using different models to serve as distinct roles, capitalizing on their respective strengths to further enhance the effectiveness of the pipeline. 
{
    \small
    \bibliographystyle{ieeenat_fullname}

    % \bibliography{egbib_org}
}

\end{document}